\documentclass[10pt, conference, letterpaper]{IEEEtran}
\usepackage{textgreek}

\usepackage[noadjust]{cite}

\ifCLASSINFOpdf
\else
\fi
\usepackage{graphicx}
\usepackage{epstopdf}
\DeclareGraphicsExtensions{.eps}
\graphicspath{ {images/} }

\usepackage{amsmath}

\usepackage{amsthm}

\newtheorem*{remark}{Remark}
\newtheorem{theorem}{Theorem}[section]
\newtheorem{lemma}[theorem]{Lemma}

\theoremstyle{definition}
\newtheorem{definition}{Definition}[section]

\begin{document}
%
\title{On the Dynamics of a Recurrent Hopfield Network}

\author{\IEEEauthorblockN{Rama Garimella}
\IEEEauthorblockA{Department of Signal Processing\\
Tampere University of Technology\\
Tampere, Finland\\
Email: rammurthy@iiit.ac.in}
\and
\IEEEauthorblockN{Berkay Kicanaoglu}
\IEEEauthorblockA{Department of Signal Processing\\
Tampere University of Technology\\
Tampere, Finland\\
Email: berkay.kicanaoglu@student.tut.fi}
\and
\IEEEauthorblockN{Moncef Gabbouj}
\IEEEauthorblockA{Department of Signal Processing\\
Tampere University of Technology\\
Tampere, Finland\\
Email: moncef.gabbouj@tut.fi\\
}}


%


\maketitle

\begin{abstract}
In this research paper novel real/complex valued recurrent Hopfield Neural Network (RHNN) is proposed. The method of synthesizing the energy landscape of such a network and the experimental investigation of dynamics of Recurrent Hopfield Network is discussed. Parallel modes of operation (other than fully parallel mode) in layered RHNN is proposed. Also, certain potential applications are proposed.
\end{abstract}


%
\IEEEpeerreviewmaketitle

\section{Introduction}
	In the research field of electrical engineering, following the principle of parsimony, 
	many natural as well as artificial systems were modeled as linear systems. However it was
	realized that many natural/artificial phenomena are inherently non-linear. For instance, 
	many models of neural circuits require non-linear difference/differential equations for 
	modeling their dynamics. Hopfield effectively succeeded in arriving at a model of associative 
	memory based on McCulloch-Pitts neuron. The discrete time Hopfield network represents a 
	non-linear dynamical system. 
			
	Hopfield Neural Network spurred lot of research efforts on modeling associative memories (such
	as the bi-directional associative memory). However, it was realized that the Hopfield neural 
	network is based on undirected graph(for topological representation of network architecture) and 
	hence does not have any directed cycles in the network architecture. Thus, in that sense Hopfield 
	Neural Network (HNN) does not constitute a recurrent neural network. Many recurrent networks, such 
	as Jordan-Elman networks were proposed and studied extensively. After literature survey, we were 
	motivated to propose an Artificial Neural Network (ANN), in the spirit of HNN, but the network 
	architecture is based on a directed graph with directed cycles in it.  
	
	Our research efforts are summarized in this paper. In Section-2, a novel real as well as complex Recurrent Hopfield 
	Network is discussed. In Section-3, experimental investigations of Recurrent Hopfield neural Network are discussed.
	In Section-4, some applications are proposed. The research paper concludes in Section-5.

\section{Recurrent Hopfield Network(RHN) }
	Recurrent  Hopfield  Neural  Network (RHNN) is  an  Artificial  Neural  Network  model. It  is  a  nonlinear  	
	dynamical  system  represented  by  a weighted,  directed  graph.  The  nodes  of  the  graph  represent  artificial  
	neurons  and  the  edge  weights  correspond  to  synaptic  weights. At  each  neuron/node,  there  is  a  
	threshold  value. Thus,  in  summary,  a Recurrent   Hopfield   neural  network  can  be  represented  by  a  
	synaptic  weight  matrix, M  that  is  not  necessarily symmetric  and  a  threshold  vector, T. The  order  of  
	the  network corresponds to the number of  neurons. Such a neural network potentially has some   
	directed cycles in the associated graph. 
	
	Every  neuron  is  in  one  of  the  two  possible  states   +1  or  -1. Thus,  the  state  space  of  Nth  order 
	Recurrent  Hopfield  Neural  Network (RHNN)  is  the symmetric  N-dimensional  unit  hypercube.  Let  the  state  
	of  \textit{i}th neuron at time t be  denoted  by 
	
	\begin{equation*}
	V_i(t) \textit{ where } V_i(t)\in \{ +1 \textit{ or}-1 \}.
	\end{equation*} 
	
	Thus, the state of the	non-linear dynamical system is represented by the N x 1 vector, \begin{math} \bar{V}(t) 
	\end{math}. The  state updation at the  \textit{i}th  node  is  governed  by  the  following  equation
	
	\begin{equation} \label{eq:2.1}
	V_i(t+1) = Sign(\sum\limits_{j=1}^N M_{ij} V_j(t) - T_i)
	\end{equation} 
	
	where  Sign (.)  is   the  Signum  function.  In  other  words,  \begin{math} V_i(t+1)\end{math}  is  +1  if  the  
	term  in  the brackets  is  non-negative,  otherwise  \begin{math} V_i(t+1)\end{math} is -1.  Depending  on  the  
	set  of  nodes at  which  the state  updation by Eq. ~\ref{eq:2.1}\  is  performed  at  any  time  \textit{t},  the  
	Recurrent	Hopfield  Neural	Network	(RHNN) operation  is  classified  into  the  following  modes.
	
	\begin{itemize}
		\item Serial  Mode: 
		The  state  updation  in Eq.~\ref{eq:2.1}\  is  performed  exactly  at  one of  the  nodes/neurons  at  	
		time \textit{t}.
		\item Fully Parallel Mode: 
		The state updation in  Eq.~\ref{eq:2.1}\  is  performed  simultaneously  at  all  the  N  nodes/neurons at
		time \textit{t}.
	\end{itemize}
	
	\begin{remark}[1]
	If  the  matrix  M  is  symmetric,  then   the  dynamics  of   Recurrent  Hopfield  Neural  Network (RHNN) is   exactly  
	same  as  the  same  as  that  of  Ordinary   Hopfield Neural  Network   (OHNN). 	
	\end{remark}

	For  the  sake  of  completeness,  we  now  include the  dynamics  of  Ordinary   Hopfield  Neural  Network (OHNN)
	with  M  being  a symmetric  matrix [1]. In  the  state  space  of Ordinary   Hopfield  Neural  Network 
	(OHNN),  a  non-linear  dynamical  system,  there  are  certain  distinguished  states, called the 
	\textit{stable states}.
	
	\theoremstyle{definition}
	\begin{definition}
	A  state, \begin{math} \bar{V}(t) \end{math} is  called    a  stable  state  if  and  only  if
	
		\begin{equation} \label{eq:2.2}
		\bar{V}(t) = Sign(M \bar{V}(t) - T) 
		\end{equation}
	
	\end{definition}
	
	Thus,  once  the  Hopfield  Neural  Network (HNN)  reaches  the  stable  state, irrespective  of  the  mode  of  
	operation  of  the  network,  the  HNN  will  remain  in  that  state  for  ever.  Thus,  there  is  no  further  
	change  of  the  state  of  HNN  once  a  stable  state  is  reached.  The  following  convergence  Theorem  
	summarizes  the  dynamics  of Ordinary Hopfield  Neural  Network  (OHNN).  It  characterizes  the  operation  of  
	neural  network  as  an  associative  memory.
	
	\begin{theorem} \label{Theorem1}
	Let  the  pair \begin{math} N = (M,T) \end{math} specify  a  Ordinary  Hopfield  Neural Network  (with  M  being  
	symmetric).Then the following  hold  true:
	
	[1]  Hopfield:  If   N  is  operating  in  a  serial  mode  and  the  elements  of  the diagonal   of  M   
	are  non-negative,  the  network  will   always  converge   to  a stable  state   (i.e.   there  are  no  
	cycles   in  the  state  space).

  [2] Goles:  If   N   is  operating  in  the  fully  parallel  mode,  the   network  will  always converge  
	to  a  stable  state  or  to  a  cycle  of  length  2 (i.e  the  cycles  in  the  state  space  are  of 
	length  \begin{math} \le 2 \end{math}).

	\end{theorem}
	
	\begin{remark}[2]
	It  should  be  noted   that  in  [4],  it  is  shown  that  the  synaptic weight matrix of OHNN can  be  assumed  
	to  be  an  arbitrary  symmetric  matrix.
	\end{remark}

\subsection{Layered   Recurrent  Hopfield   Neural   Network}
	We  now  consider  a  Recurrent Hopfield  Neural  Network  where  the  neurons  (artificial)  are  organized 
	into finitely  many  layers  (with  \begin{math} \ge 1 \end{math}  neurons   in  each  layer).  Thus  the  synaptic 
	weight   matrix  of  
	such  a  neural   network  can  be  captured  by  a  block  matrix, W which  is  not  necessarily  symmetric.   
	Traditionally,  the  convergence  Theorem associated  with  Ordinary  Hopfield  Neural  Network  (OHNN) (i.e.  
	Theorem ~\ref{Theorem1}) effectively  considered  only  (i)  Serial  Mode  and  (ii)  Fully  Parallel  Mode.  
	But the  arrangement  of  neurons  in   multiple  layers  naturally  leads  to  operation  of the  Hopfield  network  
	(Ordinary  as  well  as  Recurrent)  in  other  parallel  modes of   operation. This  is  accomplished  in  the  
	following  manner:
	
	\begin{itemize}
		\item It  should  be  noted  that  the  state  vector  of the Ordinary/Recurrent  Hopfield Neural  
		Network	at time \textit{t} can  be  partitioned  into  sub-vectors  in  the  following manner.
		
			\begin{math} 
			\bar{V}(t) = [\bar{V}^{(1)}(t): \bar{V}^{(2)}(t): ... : \bar{V}^{(L)}(t)] \\
			\end{math}
		where 
		\begin{math} \bar{V}^{(1)}(t)  \end{math} denotes   the  state  of  neurons  in  layer  1  and  L  is 
		the  number of layers.
		
		For  the  sake  of  notational  convenience,  suppose  that  all  the  'L'  layers have  the  same  number  of  
		neurons  in  it   and  state  updation  at  any  time instant takes  place  simultaneously  at  all  nodes  in  
		any  one  layer.  This  clearly  corresponds to  parallel  mode  of  operation which  is  NOT  the  FULLY  
		PARALLEL  MODE (if there  is  a  single  neuron  in  each  layer,  it  corresponds  to  the  serial  mode  of 
		operation). The  state  updation  in  such  a  'parallel  mode'  can  be  captured  through the  following  
		equation:  For \begin{math} 1\le j \le L \end{math},  we  have   that
		
		\begin{equation*}
		\bar{V}^{(j)}(t+1) = Sign(\sum\limits_{k=1}^L W_{(j,k)} \bar{V}^{(k)}(t) - \bar{T}_j)
		\end{equation*} 
		
		where \begin{math} W_{(j,k)} \end{math} is  the  sub-matrix  of  W  and  \begin{math} \bar{T}_j \end{math} is 
		the  corresponding threshold vector. It  can  be  shown  that  there  is  no  loss  of 	
		generality in  assuming  that   the  threshold  vector  is  a  zero  vector.
		
		\textbf{Note: } To  the  best  of  our  knowledge,  the  dynamics  of  even  OHNN  in other  parallel  modes  
		of  operation  (not  fully  parallel)  has  not  been  fully investigated.
		
		\begin{remark}[3]
		This  idea  of  layering  the  neurons  enables one  to  study  the   dynamics of   Recurrent  Hopfield  Neural  	
		Network   with  various  types  of  directed cycles  in  it. Theoretical  efforts  to  capture  the  dynamics  of  
		such  neural networks  are  underway.  In  the Section-3,  we  summarize  our experimental  efforts.
		\end{remark}
		
		\item \textbf{Energy  Landscape Synthesis:} \\
		Now  we  investigate  the  energy  landscape  visited by  a   Recurrent  Hopfield  Neural  Network (RHNN) and  	
		relate  it  to  the  energy landscape visited  by  the  associated  Ordinary  Hopfield  Neural  Network (OHNN).
		\item Consider  the quadratic energy  function  associated  with  the  dynamics  of  a  RHNN  whose synaptic weight 
		matrix  is  the   non-symmetric  matrix, W. It  is  easy  to  see  that
		
		\begin{align*}
		\bar{V}^T(n)W\bar{V}(n) &= \bar{V}^T(n) \left(\frac{W + W^T}{2}\right) \bar{V}(n)\\
	  &= \bar{V}^T(n) \bar{W} \bar{V}(n)
		\end{align*}
		
		where \begin{math} \bar{W} \end{math} is  the  symmetric  matrix  associated  with  W, i.e. symmetric  part.
		
		\item Now,  from  the  above  equation,  it  is  clear  that  the  energy  landscape  visited by  the  
		RHNN  with  synaptic  weight  matrix W is  exactly  same  as  the  energy landscape  visited  by  the  
		OHNN  with  the  symmetric  synaptic  weight  matrix \begin{math}\bar{W}\end{math}. Now,  we  synthesize 
		the  energy  landscape of  RHNN (or  associated  OHNN) in  the   following  manner: \\
		We  need  the  following  definition  in  the  succeeding  discussion.
	\end{itemize}
	\theoremstyle{definition}
	\begin{definition}
	The  local  minimum  vectors  of  the energy  function  associated  with  W  are  called \textbf{anti-stable 
	states}. Thus,  if   u  is   an   anti-stable  state  of   matrix  W,  then  it  satisfies  the  condition
	
		\begin{equation*}
		u= -Sign(Wu).
		\end{equation*}
	
	\end{definition}
	Using  the  definitions,  we  have  the  following   general  result.
	
	\begin{lemma}
	If  a  corner  of  unit   hypercube  is  an  eigenvector  of   W  corresponding   to   positive/negative  	
	eigenvalue,  then   it  is  also  a  stable/anti-stable   state.
	\end{lemma}
	
	\begin{IEEEproof}
	Follows  from  the  utilization  of  definitions  of  eigenvectors,  stable/anti-stable  states. Q.E.D.
	\begin{itemize}
	\item \textbf{Synthesis  of  Symmetric  Synaptic  Weight  Matrix  with desired  Energy  Landscape:} \\
	Let \begin{math} \lbrace \frac{\mu_j}{N}\rbrace \end{math} where \begin{math} j \in \lbrace 0, 1, ...,S \rbrace
	\end{math} be    desired   positive   eigenvalues (with \begin{math} \mu_j \end{math} being  the  desired   
	stable  value)  and  let \begin{math} \lbrace X_j \rbrace (j \in \lbrace 0, 1, ...,S \rbrace)\end{math}
	be  the  desired   stable  states  of  \begin{math} \bar{W} \end{math}. Then   it  is  easy  to  see  that   the   	
	following  symmetric matrix   constitutes  the   desired   synaptic   weight   matrix  of  Ordinary Hopfield Neural  
	Network (Using the  spectral  representation  of   symmetric  matrix \begin{math} \bar{W} \end{math}):
		
		\begin{equation*}
		\bar{W} = \sum\limits_{j=1}^S\mu_j X_j X_j^T
		\end{equation*}
	It  is clear  that  the  synaptic  weight  matrix  is  positive  definite.
	In  the  same  spirit  as  above,  we  now   synthesize  a  synaptic  weight  matrix,  with  desired  stable/anti-
	stable  values  and   the  corresponding   stable / anti-stable  states.   Let  \begin{math}\lbrace X_j \rbrace 
	\end{math} where \begin{math} j \in \lbrace 0, 1, ...,S \rbrace	\end{math} be  desired	orthogonal   stable  states   
	and  \begin{math}\lbrace Y_j \rbrace \end{math} where \begin{math} j \in \lbrace 0, 1, ...,L \rbrace	\end{math}  be  
	the  desired  orthogonal  anti-stable  states.  Let  the desired  stable  states  be  eigenvectors   corresponding   
	to  positive  eigenvalues  and  let  the  desired  anti-stable  states  be eigenvectors   corresponding   to  
	negative  eigenvalues.  The   spectral  representation  of  desired   synaptic weight  matrix  is  given  by 
		\begin{equation*}
		W = \sum\limits_{j=1}^S\frac{\mu_j}{N} X_j X_j^T - \sum\limits_{j=1}^L\frac{\beta_j}{N} Y_j Y_j^T
		\end{equation*}
	where  \begin{math}\mu_j\end{math}'s are  desired   positive  eigenvalues   and  \begin{math}-\beta_j\end{math}'s   
	are   desired  negative  eigenvalues. Hence  the  above   construction   provides  a  method  of  arriving  at  
	desired   energy  landscape (with  orthogonal   stable/anti-stable   states   and   the  corresponding   positive/negative  energy  values).
	
	\end{itemize}
	\end{IEEEproof}
	
	\begin{remark}[4]
  It  can  be  easily  shown  that  Trace(W)  is  constant  contribution (DC  value)  to  the  value  of  quadratic  
	form  \begin{math}X^T WX\end{math}  at  all  corners,  X  of  unit  hypercube.  Thus,  the   location  of  stable/
	anti-stable  states is  invariant  under  modification  of  Trace(W)  value.  Thus,  from  the  standpoint   of   
	location/computation of stable/anti-stable  states,  Trace(W)  can  be  set  to  zero.  
	\end{remark}
	
	\begin{itemize}
	\item 	It  should  be  kept  in  mind  that  the   dynamics  of  OHNN  with  the  symmetric  synaptic  weight  
	matrix  \begin{math} \bar{W} \end{math} is   summarized  by  Theorem \ref{Theorem1},  i.e.  either   there is
	convergence  to  stable  state  or  cycle  of  length  2  is  reached.  But  in the  case
	RHNN  with  non-symmetric  synaptic  weight  matrix, there  may  be  no  convergence
	or  the  cycles  are  of  length  strictly  larger  than  2.  In  the Section-3,
	we  summarize  our  experimental  efforts.
	\end{itemize}

\subsection{Novel Complex Recurrent Hopfield Neural Network}
In  research   literature  on artificial  neural  networks,  there  were  attempts  by  many   researchers  to  propose  Complex Valued  Neural  Networks (CVNNs)  in which  the   synaptic  weights,  inputs  are  necessarily  complex  numbers [2, MGZ].  In  our  research  efforts   on  CVNNs,  we   proposed  and  studied  one  possible  Ordinary  Complex  Hopfield  Neural  Network  (OCHNN)  first  discussed  in   [3, RaP], [5-12].  The  detailed  descriptions  of  such  an  artificial  neural  network   requires  the  following  concepts.	
\begin{enumerate}
	\item \textit{Complex Hypercube}:  Consider  a  vector  of   dimension 'N'  whose  components  assume   values  in  
	the following  set   \begin{math} \lbrace 1 + j1 ,  1 - j1, -1 + j1, -1 - j1 \rbrace\end{math}.Thus  there  are   
	points  as the  corners  of  a  set called the  ‘Complex  Hypercube”.
	\item \textit{Complex Signum Function}: Consider  a  complex  number  'a + jb'.  The  'Complex  Signum  Function'  	
	is defined  as  follows:
		\begin{equation*}
		Csign (a + \textit{j}b) = Sign(a) + \textit{j}Sign(b)
		\end{equation*}
\end{enumerate}
In  the  same  spirit  of   the   real  valued  RHNN, we  briefly  summarize  Complex  Recurrent  Hopfield   Neural  Network (CRHNN):
\begin{itemize}
\item The  complex  synaptic  weight  matrix  of  such  an  artificial  neural  network (CRHNN)  need  NOT be   a  
	Hermitian  matrix.  The   activation  function  in  such  a  complex  valued   neural  network  is  the  complex  		
	signum  function.   The   convergence  Theorem  associated  with  Ordinary  Complex  Hopfield  Neural  Network 
	(OCHNN) is  provided  in [3, RaP].  We  are  experimentally  investigating  the  dynamics  of   CRHNN.
\end{itemize}

	\section{Dynamics of Recurrent Hopfield Network:\\\textit{Experiments}}
	In this section, certain network configurations are simulated with various initial conditions (lying on the unit 
	hypercube) in order to provide insight on the dynamics of the Recurrent Hopfield Networks.
	\subsection{Topology of the networks}
	The recurrent neural networks whose dynamics are presented in this study have asymmetric synaptic weight 
	matrices with zeros on the diagonal (i.e. no self loops are allowed. As discussed in \textit{Remark(4)}, this assumption
	can be made without loss of generality). Edge weights (the synaptic weights) are allowed 
	to assume both positive and negative signs. The size of the network (i.e. number of neurons) directly determines 
	the	size of synaptic weight matrix. In the experiments, we are going to investigate three cases: synaptic weight 
	matrices with	(i) \textit{both positive and negative}, (ii) \textit{only positive} and (iii) \textit{only negative} 
	entries.
	\subsection{Initialization of the network}
	A Recurrent Hopfield Network, in general, can be randomly initialized in a state \textit{s} where
	
	\begin{equation*}
	\textit{s}\in \Psi.
	\end{equation*}
	
	\begin{math} \Psi \end{math} is the state space and 
	
	\begin{equation*}\label{eq:stateSpaceSize}
	\left|\Psi\right| = 2^n. 
  \end{equation*}
	The size of the state space exponentially increases as the number of neurons are increased. In order to reduce the 
	computational overhead, the number of initial states utilized are limited to a maximum of 1024. 
	\begin{itemize}
	\item The decimals numbers in the range [0,1024) are converted to binary and then all 0's are mapped to -1's 
	(From this point on, states are going to be referred by using their decimal values).
	\end{itemize}
	\subsection{Modes of operation in experiments}
	During the experiments, both parallel and serial modes are tested. However, serial mode results in two specific 
	cases. The first is the case where network converges and the second is when there is a cycle of 2 as iterations 
	proceed. On the other hand, parallel mode exhibits more interesting outcomes. Hence, the focus during the experiments
	have been placed on this update mode.  
	\subsection{Results}
	\subsubsection{Convergence and cycles}
	Networks with various polarities of synaptic connections behave expectedly in different manners. The experiments 
	have led to certain types of observations in terms of convergence and cycle characteristics. In vast majority of 
	trials, it has been observed that given an initial state from \begin{math} \Psi \end{math}, the network either 	
	converges to a single state or converges into a certain pattern of cycle in multiple iterations. \\
	
	\begin{figure}[!t]
		\centering 
			\[ \left( \begin{array}{ccccccc}
			0	& 11 & 13	& -15 &	6	& -8 & -12\\
		  -20	& 0 &	5	& -5 & 4 & -16 & 15\\
			5	& -17 & 0	& 7	& 13 & 13	& -6\\
		  10 & 8 & -9	& 0	& 1 & -19	& 10\\
			4	& 19 & 17 &-3	& 0	& 5	& -8\\
		 -11 & -20 & 9 & 5 & -7 & 0 & 16\\
		 -12 & 4 & 2 & 1 & -17 & 13	& 0\\
		             \end{array} \right)\] 	
		\caption{Toy synaptic weight matrix}
		\label{fig:toyMatrix}
	\end{figure}
	
	\begin{itemize}
	\item In networks with both positive and negative synaptic connections, depending on the network size, convergence 	
	can take place immediately if the initial state is located in a small proximity to one of the basins of attraction in 
	the energy landscape. However simple generalization of the network dynamics is not possible. The network can be led to 
	certain cycles which are of various lengths. One main observation is that a collection of distinct initial states can 
	be driven into one or more different cycles. This result is interesting since it may imply that a Recurrent Hopfield 
	Network can be used as a function to map signals into certain common cycles. To illustrate the possible state 
	transitions, the network in \figurename \ref{fig:toyMatrix} is to be used.
	\end{itemize}

	\begin{itemize}
	\item When parallel mode of update is applied on this network, the results revealed that there exist 6 cycles in 
	which the RHNN loops given all possible initializations. In this mode, length of the cycles varied between 2 and 4 
	while in the serial mode of operation, the network reached cycles of length of at most two. Furthermore, it often 
	converged to certain minima points in state space. To note, the cycles of length 1 are used to refer to the stable 
	states. The observed cycles, following the naming convention provided previously, are listed in \tablename 
	~\ref{cycles1}.
	\item If the non-symmetric synaptic weight matrix is non-negative or non-positive, it is observed that the network 
	converges to a single state or to a cycle of length 2.
	\item If the non-symmetric synaptic weight matrix has positive and negative elements, it is observed that the cycles 	
	are of length strictly equal or larger than 2.
	\item It is observed that in the case of various RHNNs, there are more than one directed cycles in the state space (reached with a change of initial conditions).
	\item In some special cases a single cycle is reached.
	\end{itemize}
	\begin{itemize}
	\item The simulations show that the networks of mixed-polarity connections are favoring certain states or 
	combinations of	the whole possible state space. The parallel mode yielded that inputs that are close in terms of 
	Hamming distance usually go through the same cycles or end up at certain stable states. 
	\item Larger networks with more than 20 neurons are observed to fail to reach at detectable patterns or single state 
	in the parallel mode within reasonable number of iterations.	
	\end{itemize}
	
	Next section will focus on the 
	energy values associated with various states. The energy values may be useful to make better conclusions on the 
	observations. 
	
	
	\begin{table}[!t]
	\renewcommand{\arraystretch}{1.3}
	\caption{The cycles for the toy example}
	\label{cycles1}
	\centering
		\begin{tabular}{ c || c }
		\hline
		\bfseries Mode & \bfseries Cycles \\
		\hline\hline
		Parallel & 35, 58-69, 42-5, 21-106, 22-119-92-94, 85-122 \\ \hline
		Serial & 11, 35, 36, 41, 45, 57, 61, 63, 64, 66, 70, 86,  \\
		& 91, 92, 100-12, 27-115, 118-112, 118-31 \\
		\hline
		\end{tabular}
	\end{table}	

	\subsection{Energy of the Network during cycles and convergence}
\begin{figure}[!t]
	\centering
	\includegraphics[width=2.5in]{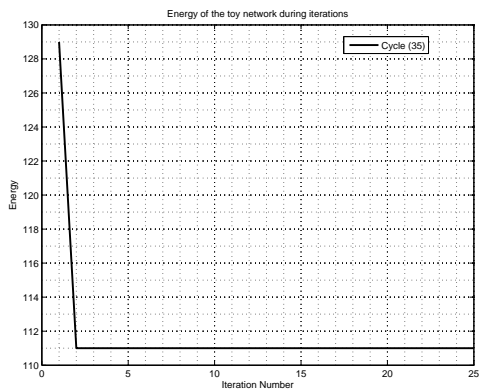}
	\caption{Energy of the network during the cycle 35.}
	\label{fig:energy1}
	\end{figure}

\begin{figure}[!t]
	\centering
	\includegraphics[width=2.5in]{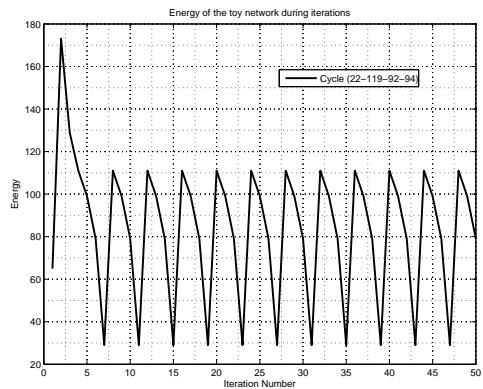}
	\caption{Energy of the network during the cycle 22-119-92-94.}
	\label{fig:energy5}
	\end{figure}

	The Recurrent Hopfield Networks experimented in this study exhibit two characteristics. First characteristic is that 
	the network reaches a constant energy value when it gets trapped in a single state or a set of states (a cycle). The
	second characteristic is that the energy of the network fluctuates around a certain DC value. The latter has been 
	observed to occur in cycles only. The typical energy states of the toy example during 128 trials are given in figures 
	\figurename \ref{fig:energy1} - \figurename \ref{fig:energy5}. It is useful to notice that the network's energy states
	(during convergence or cyclic period) which are caused by a collection of	distinct initial states are the same
	recapitulating what has been stated previously. 
	
\begin{figure}[!t]
	\centering
	\includegraphics[width=2.5in]{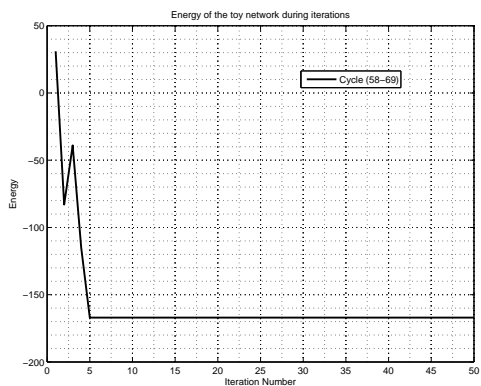}
	\caption{Energy of the network during the cycle 58-69.}
	\label{fig:energy2}
	\end{figure}

\begin{figure}[!t]
	\centering
	\includegraphics[width=2.5in]{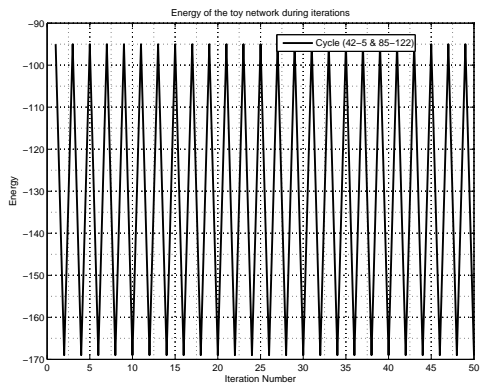}
	\caption{Energy of the network during the cycle 42-5 and 85-122.}
	\label{fig:energy3}
	\end{figure}

\begin{figure}[!t]
	\centering
	\includegraphics[width=2.5in]{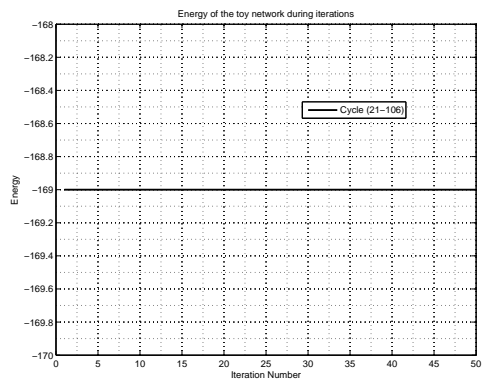}
	\caption{Energy of the network during the cycle 21-106.}
	\label{fig:energy4}
	\end{figure}

	\begin{itemize}
	\item In \figurename \ref{fig:energy1}, a common energy profile of network is shown. Usually, in the toy example, 
	the network configures itself into one of the low energy regions.
	\item \figurename \ref{fig:energy2} and	\figurename \ref{fig:energy4} shows
	that the network obtains a steady energy even during the cycles while \figurename \ref{fig:energy3} and \figurename
	\ref{fig:energy5} indicates that the network's energy oscillates as well as the state it assumes on each iteration.
	\item The observations indicate that larger networks exhibit similar behavior in the parallel mode but 
	with the exception that energy fluctuations are more often present during cyclic period. 
	\item In serial mode of operation, the energy of the network remains same if it undergoes a single state convergence 
	whereas it fluctuates if in a cycle of 2.

	\end{itemize}
		
	\section{Applications}
	Non-linear dynamical systems exhibit interesting dynamic behavior such as (i) \textit{convergence}, 
	(ii) \textit{oscillations} (periodic behavior) and (iii) \textit{chaos}.

	Researchers have identified certain applications (such as cryptography) for interesting non-linear dynamical systems. 
	Specifically neural circuits based on non-linear dynamical systems were identified with certain functional 
	capabilities of biological brains. The authors realized that Recurrent Hopfield Network in this research paper could 
	be capitalized for various applications. 
	
	\begin{itemize}
	\item The main observation is that the cycles of various lengths in the state space could be 
	utilized for applications such as "MULTI-PATTERN" based associative memory. Based on the length of cycle, a memory 
	utilizing multiple patterns (corners of the hypercube) can be synthesized.
	\item In the RHNNs investigated, sometimes CONSENSUS to a single cycle is achieved. Thus, such networks
	could be utilized for memorizing a collection of states corresponding to a cycle.
	\item In some RHNNs simulated, multiple cycles are reached when the initial condition is varied. Thus, in this case
	CONSENSUS to multiple memory groups (of states) is achieved.
	\end{itemize}

\section{Conclusion}
In this research paper, our efforts have focused on introducing a novel real/complex valued 
Recurrent Hopfield Neural Network. In addition, we have shown a method to synthesize a desired 
energy landscape for such networks. In the experimental section, the dynamics of such real valued 
Recurrent Hopfield Networks have been investigated in terms of the state transitions and energy. 
Finally, certain potential applications which exploit the shown properties are proposed.


%
%



%

\section*{References}
\begin{enumerate}[\IEEEsetlabelwidth{12}]
\renewcommand\labelenumi{[\theenumi]} 
\item \label{1} J.J. Hopfield, ``Neural  Networks  and  Physical  Systems  with  Emergent Collective  Computational  Abilities,'' Proceedings  of  National  Academy  Of  Sciences, USA Vol. 79, pp. 2554-2558, 1982. \\
\item \label{2} Mehmet Kerem Muezzinoglu, Cuneyt Guzelis and Jacek M. Zurada, ``A new design method for the Complex-valued Multistate Hopfield Associative Memory,'' IEEE Transactions on Neural Networks, Vol. 14, No. 4, July 2003.\\
\item \label{3} G. Rama Murthy and D. Praveen, ``Complex-valued Neural Associative Memory on the Complex hypercube,'' IEEE Conference on Cybernetics and Intelligent Systems, Singapore, 1-3 December, 2004.\\
\item \label{4} G. Rama  Murthy, ``Optimal  Signal  Design  for  Magnetic  and Optical  Recording  Channels,'' Bellcore  Technical  Memorandum, TM-NWT-018026,  April  1st , 1991.\\
\item \label{5} G. Rama  Murthy  and  B. Nischal, ``Hopfield-Amari Neural Network : Minimization of Quadratic forms,''  The 6th International Conference on Soft Computing and  Intelligent  Systems, Kobe Convention Center (Kobe Portopia Hotel), November 20-24, 2012, Kobe, Japan.\\
\item \label{6} Akira Hirose, ``Complex Valued Neural Networks: Theories and Applications,'' World scientific publishing Co, November 2003.\\
\newpage 
\item \label{7} G. Rama Murthy and D. Praveen, ``A Novel Associative Memory on the Complex Hypercube Lattice,'' 16th European Symposium on Artificial Neural Networks, April 2008.\\
\item \label{8} G. Rama Murthy, ``Some Novel Real/ Complex Valued Neural Network Models,'' Advances in Soft Computing, Springer Series on Computational Intelligence: Theory and Applications, Proceedings of 9th Fuzzy days, September 18-20 2006, Dortmund, Germany.\\
\item \label{9} G. Jagadeesh, D. Praveen and G. Rama Murthy, ``Heteroassociative Memories on the Complex Hypercube,'' Proceedings of 20th IJCAI Workshop on Complex Valued Neural Networks, January 6-12 2007.\\
\item \label{10} V. Sree Hari Rao and G. Rama Murthy, ``Global Dynamics of a class of Complex Valued Neural Networks,'' Special issue on CVNNS of International Journal of Neural Systems, April 2008.\\
\item \label{11} G. Rama Murthy, ``Infinite Population, Complex Valued State Neural Network on the Complex Hypercube,'' Proceedings of International Conference on Cognitive Science, 2004.\\
\item \label{12} G. Rama Murthy, ``Multidimensional Neural Networks-Unified Theory,'' New Age International Publishers, New Delhi, 2007.
\end{enumerate}

%
%
%

\end{document}